\documentclass[authoryear,12pt]{elsarticle}

\usepackage{times}
\usepackage{latexsym}
\usepackage{amsmath,amssymb}
\usepackage{multirow}
\usepackage{url}
\usepackage{graphicx}

\journal{arXiv}

\begin{document}

\begin{frontmatter}

%% Title, authors and addresses

%% use the tnoteref command within \title for footnotes;
%% use the tnotetext command for the associated footnote;
%% use the fnref command within \author or \address for footnotes;
%% use the fntext command for the associated footnote;
%% use the corref command within \author for corresponding author footnotes;
%% use the cortext command for the associated footnote;
%% use the ead command for the email address,
%% and the form \ead[url] for the home page:
%%
%% \title{Title\tnoteref{label1}}
%% \tnotetext[label1]{}
%% \author{Name\corref{cor1}\fnref{label2}}
%% \ead{email address}
%% \ead[url]{home page}
%% \fntext[label2]{}
%% \cortext[cor1]{}
%% \address{Address\fnref{label3}}
%% \fntext[label3]{}

\title{Statistical sentiment analysis performance in Opinum}

%% use optional labels to link authors explicitly to addresses:
%% \author[label1,label2]{<author name>}
%% \address[label1]{<address>}
%% \address[label2]{<address>}

\author{Boyan Bonev, Gema Ram\'irez-S\'anchez, Sergio Ortiz Rojas}

\address{  Prompsit Language Engineering \\
  Avenida Universidad, s/n. Edificio Quorum III.\\
  03202 Elche, Alicante (Spain) \\
  {\tt \{boyan,gramirez,sortiz\}@prompsit.com} 
}

\begin{abstract}
The classification of opinion texts in positive and
negative is becoming a subject of great interest in sentiment analysis.
The existence of many labeled opinions motivates the use of
statistical and machine-learning methods. First-order statistics
have proven to be very limited in this field.
The Opinum approach is based on the order of the words
without using any syntactic and semantic information.
It consists of building one  
probabilistic model for the positive and another one for the negative opinions.
Then the test opinions are compared to both models and a decision
and confidence measure are calculated.
In order to reduce the complexity
of the training corpus we first lemmatize the texts and
we replace most named-entities with wildcards.
Opinum presents an accuracy above 81\% for Spanish opinions in
the financial products domain. In this work we discuss which are
the most important factors that have an impact on the classification performance.

\end{abstract}

\begin{keyword}
%% keywords here, in the form: keyword \sep keyword
sentiment analysis \sep opinion classification \sep language model
%% MSC codes here, in the form: \MSC code \sep code
%% or \MSC[2008] code \sep code (2000 is the default)

\end{keyword}

\end{frontmatter}

% \linenumbers

%% main text
\section{Introduction}

Most of the texts written by humans reflect some kind of sentiment.
The interpretation of these sentiments depend on the linguistic skills 
and emotional intelligence of both the author and the reader, but
above all, this interpretation is subjective to the reader.
They don't really exist in a string of
characters, for they are subjective states of mind. 
Therefore sentiment analysis is a prediction of how 
most readers would react to a given text.

There are texts which intend to be objective and texts which
are intentionally subjective. The latter is the case of opinion texts,
in which the authors intentionally use an appropriate language to express their
positive or negative sentiments about something. In this paper we
work on the classification of opinions in two classes: those expressing positive sentiment
(the author is in favour of something) and those expressing negative sentiment,
and we will refer to them as positive opinions and negative opinions.

Sentiment analysis is possible thanks to the opinions available on-line.
There are vast amounts of text in fora, user reviews, comments in blogs
and social networks. It is valuable for
marketing and sociological studies to analyse these freely available
data on some definite subject or entity.
Some of the texts available do include opinion information like
stars, or recommend-or-not, but most of them do not. A good corpus
for building sentiment analysis systems would be a set of opinions
separated by domains. It should include some information about the 
cultural origin of authors and their job, and each opinion should
be sentiment-evaluated not only by its own author, but by many other
readers as well. It would also be good to have a marking of the subjective
and objective parts of the text. Unfortunately this kind of corpora are
not available at the moment.

In the present work we place our attention at the supervised classification of 
opinions in positive and negative. Our system, which we call 
\textit{Opinum}\footnote{An Opinum installation can be tested from a web interface at \url{http://aplica.prompsit.com/en/opinum}}, is trained from
a corpus labeled with a value indicating whether an opinion is positive or negative.
The corpus was crawled from the web and it consists of a 160MB
collection of Spanish opinions about financial products. Opinum's approach
is general enough and it
is not limited to this corpus nor to the financial domain.

There are state-of-the-art works on sentiment analysis which care about differentiating
between the objective and the subjective part of a text. For instance,
in the review of a film there is an objective part and then the opinion (\cite{Raaijmakers2008}).  
In our case we work directly with opinion texts and we do not make such difference.
We have noticed that in customer reviews, even when stating objective facts, some positive or
negative sentiment is usually expressed.

Many works in the literature of sentiment analysis take lexicon-based
approaches, like \cite{Taboada2011}. For instance \cite{Hu2004,Google:08} use WordNet to extend
the relation of positive and negative words to other related lexical units.
However the combination of which words appear together may also be important
and there are comparisons of different Machine learning approaches (\cite{Pang2002})
in the literature, like Support Vector Machines, k-Nearest Neighbours, Naive-Bayes, 
and other classifiers based on global features.
In the work of \cite{Mcdonald2007}, structured models are used to infer the sentiment from 
different levels of granularity. They score cliques of text based on a high-dimensional
feature vector.

In the Opinum approach we score each sentence based on its
$n$-gram probabilities. For a complete opinion we sum the scores of all its sentences.
Thus, if an opinion has several positive sentences and it finally concludes with a
negative sentence which settles the whole opinion as negative, Opinum would probably fail.
The $n$-gram sequences are good at capturing phrasemes (multiwords), the motivation for which
is stated in Section~\ref{sec:hypothesis}. Basically, there are phrasemes
which bear sentiment. They may be different depending on the domain and it is
recommendable to build the models with opinions belonging to the target domain,
for instance, financial products, computers, airlines, etc.
A study of domain adaptation for sentiment analysis is presented in the work of \cite{Blitzer2007}.
In Opinum different classifiers would be built for different domains. Building
the models does not require the aid of experts, only a labeled set of opinions is
necessary. 
Another contribution of Opinum is that it applies some simplifications on the original text of
the opinions for improving the performance of the models. 

In the remainder of the paper we first state the
motivation of our approach in Section~\ref{sec:hypothesis},
then in Section~\ref{sec:approach} we describe in detail the Opinum approach.
In Section~\ref{sec:experiments} we present our experiments with
Spanish financial opinions. In Section~\ref{sec:discussion} we discuss which are
the most important factors that have an effect on the classification performance.
Finally we state some conclusions and future work in Section~\ref{sec:conclusions}.

\section{Hypothesis\label{sec:hypothesis}} 
When humans read an opinion, even if they do not understand it completely
because of the technical details or domain-specific terminology, in most cases they can
notice whether it is positive or negative. 
The reason for this is that the author of the opinion,
consciously or not, uses nuances and structures which show a positive or negative
feeling. Usually, when a user writes an opinion about a product, the 
intention is to communicate that subjective feeling, apart from describing
the experience with the product and giving some technical details.

The hypothesis underlying the traditional keyword or lexicon-based
approaches (\cite{Google:08,Hu2004})
consist in looking for some specific positive or negative words. For instance,
``great'' should be positive and ``disgusting'' should be negative. Of course there are
some exceptions like ``not great'', and some approaches detect negation to invert the
meaning of the word. More elaborate cases are constructions like ``an offer you can't refuse'' or
``the best way to lose your money''. 

There are
domains in which the authors of the opinions might not use these explicit keywords. 
In the financial domain we can notice that many of the opinions which express the 
author's insecurity are actually negative, even though the words are mostly neutral. 
For example, ``I am not sure if I would get a loan from this bank'' has a
negative meaning. Another difficulty is that the same words could be positive or
negative depending on other words of the sentence: ``A loan with high interests''
is negative while ``A savings account with high interests'' is positive.
In general more complex products have more complex and subtle opinions. The opinion
about a cuddly toy would contain many keywords and would be much more explicit 
than the opinion about the conditions of a loan. Even so, the human readers can get
the positive or negative feeling at a glance.

The hypothesis of our approach is that it is possible to classify
opinions in negative and positive based on canonical (lemmatized) word sequences.
Given a set of positive opinions $\mathbf{O}^p$ and a set of negative opinions
$\mathbf{O}^n$, the probability distributions of their $n$-gram word
sequences are different and can be compared to the $n$-grams of a new
opinion in order to classify it. 
In terms of statistical language models, given the language models $\mathit{M}^p$ and
$\mathit{M}^n$ obtained from  $\mathbf{O}^p$ and $\mathbf{O}^n$, 
the probability $p^p_o = P(o|\mathbf{O}^p)$ that a new opinion would be
generated by the positive model is smaller or greater than the
probability $p^n_o = P(o|\mathbf{O}^N)$ that a new opinion would be
generated by the negative model. 

We build the models based on sequences of canonical words in order to
simplify the text, as explained in the following section. We also
replace some named entities like names of banks, organizations and people
by wildcards so that the models do not depend on specific entities.

\section{The Opinum approach\label{sec:approach}}

The proposed approach is based on $n$-gram language models. Therefore
building a consistent model is the key for its success. In the field of 
machine translation a corpus with size of 500MB
is usually enough for building a $5$-gram language model, depending on the
morphological complexity of the language.

In the field of sentiment analysis it is very difficult to find a big corpus
of context-specific opinions. Opinions labeled with stars or a positive/negative
label can be automatically downloaded from different customers' opinion websites.
The sizes of the corpora collected that way range between 1MB and 20MB for both
positive and negative opinions.

Such a small amount of text would be suitable for bigrams and would capture
the difference between ``not good'' and ``really good'', but this is not enough
for longer sequences like ``offer you can't refuse''. In order to build 
consistent $5$-gram language models we need to simplify the language complexity
by removing all the morphology and replacing the surface forms by their
canonical forms. Therefore we make no difference between 
``offer you can't refuse'' and ``offers you couldn't refuse''.

We also replace named entities by wildcards: \textit{person\_entity},
\textit{organization\_entity} and \textit{company\_entity}. Although these replacements
also simplify the language models to some extent, their actual purpose is to 
avoid some negative constructions to be associated to concrete entities. For instance,
we do not care that ``do not trust John Doe Bank'' is negative, instead we
prefer to know that ``do not trust company\_entity'' is negative regardless of the
entity. This generality allows us to better evaluate opinions about new entities.
Also, in the cases when all the opinions about some entity E1 are good and all the opinions
about some other entity E2 are bad, entity replacement prevents the models from
acquiring this kind of bias.

Following we detail the lemmatization process, the named entities detection 
and how we build and evaluate the positive and negative language models.

\subsection{Lemmatization\label{subsec:lemmatization}}

Working with the words in their canonical form is for the sake of
generality and simplification of the language model.
Removing the morphological information does not change the
semantics of most phrasemes (or multiwords). 

There are some
lexical forms for which we keep the surface form or we add
some morphological information to the token. These exceptions
are the subject pronouns, the object pronouns and the possessive forms.
The reason for this is that for some phrasemes the personal information
is the key for deciding the positive or negative sense. For instance,
let us suppose that some opinions contain the sequences 
\begin{eqnarray}
o_t&=& \text{``They made money from me''},\nonumber \\
o_i&=& \text{``I made money from them''}.\nonumber  
\end{eqnarray}
Their lemmatization, referred to as
$\mathcal{L}_0(\cdot)$, would be\footnote{The 
notation we use here is for the sake of readability and it
slightly differs from the one we use in Opinum.}
\begin{eqnarray}
\mathcal{L}_0(o_t)=\mathcal{L}_0(o_i)=\text{``SubjectPronoun make money} \nonumber \\
                   \text{ from ObjectPronoun''},\nonumber 
\end{eqnarray}
Therefore we would have equally probable $P(o_t|M^p) = P(o_i|M^p)$
and $P(o_t|M^n) = P(o_i|M^n)$, which does not
express the actual sentiment of the phrasemes. In order to capture
this kind of differences we prefer to have 
%$\mathcal{L}_1(o_t)=$ ``subjectpronoun\_3p make money from objectpronoun\_1p'' and
%$\mathcal{L}_1(o_i)=$ ``subjectpronoun\_1p make money from objectpronoun\_3p''.
\begin{eqnarray}
\mathcal{L}_1(o_t)&=& \text{``SubjectPronoun\_3p make money} \nonumber \\
                  && \text{ from ObjectPronoun\_1p''},\nonumber \\
\mathcal{L}_1(o_i)&=& \text{``SubjectPronoun\_1p make money} \nonumber \\
                  && \text{ from ObjectPronoun\_3p''}\nonumber .
\end{eqnarray}
The probabilities still depend on how many times do these lexical
sequences appear in opinions labeled as positive or negative, but with
$\mathcal{L}_1(\cdot)$ we would have that 
\begin{eqnarray}
P(o_t|M^p) < P(o_i|M^p), \nonumber \\
P(o_t|M^n) > P(o_i|M^n), \nonumber 
\end{eqnarray}
that is, $o_i$ fits better the positive model than $o_t$ does, 
and vice versa for the negative model.

In our implementation lemmatization is performed with Apertium,
which is an open-source rule-based machine translation engine. Thanks to its
modularized architecture (described in \cite{Tyers2010}) 
we use its morphological analyser and its
part-of-speech disambiguation module in order to take one lexical form as the most probable one,
in case there are several possibilities for a given surface. Apertium currently has
morphological analysers for 30 languages (most of them European), which allows us
to adapt Opinum to other languages without much effort.

\subsection{Named entities replacement}

The corpora with labeled opinions are usually limited to a number
of enterprises and organizations. For a generalization purpose we 
make the texts independent of concrete entities. We do make
a difference between names of places, people and organizations/companies. 
We also detect dates, phone numbers, e-mails and URL/IP.
We substitute them all by different wildcards.
All the rest of the numbers are substituted by a ``Num'' wildcard.
For instance, the following subsequence would have a $\mathcal{L}_2(o_e)$
lemmatization + named entity substitution:
\begin{eqnarray}
o_e = &\text{``Joe bought 300 shares}  \nonumber \\
      &\text{ of Acme Corp. in 2012''} \nonumber \\
\mathcal{L}_2(o_e) = &\text{``Person buy Num share}  \nonumber \\
      &\text{ of Company in Date''} \nonumber 
\end{eqnarray}

The named entity recognition task is integrated within the lemmatization
process. We collected a list of names of people, places, companies and organizations
to complete the morphological 
dictionary of Apertium. The morphological analysis module is still very fast, as the
dictionary is first compiled and transformed to the minimal deterministic finite automaton.
For the dates, phone numbers, e-mails, IP and URL we use
regular expressions which are also supported by the same Apertium module.

Regarding the list of named entities,
for a given language (Spanish in our experiments) we download
its Wikipedia database which is a freely available resource. We
heuristically search it for organizations, companies, places and people.
Based on the number of references a given entity has in Wikipedia's 
articles, we keep the first 1.500.000 most relevant entities, which
cover the entities with 4 references or more (the
popular entities are referenced from tens to thousands of times).

Finally, unknown surface forms are replaced by the ``Unknown'' lemma (the
known lemmas are lowercase).
These would usually correspond to strange names of products,
erroneous words and finally to words which are not
covered by the monolingual dictionary of Apertium. Therefore
our approach is suitable for opinions written in a rather correct language.
If unknown surfaces were not replaced, the frequently misspelled 
words would not be excluded, which is useful in some domains.
This is at the cost of increasing the complexity of the model, as all 
misspelled words would be included. Alternatively, the frequently
misspelled words could be added to the dictionary.

\subsection{Language models}

The language models we build are based on $n$-gram word sequences.
They model the likelihood of a word $w_i$ given the sequence of $n-1$
previous words, $P(w_i|w_{i-(n-1)},\ldots,w_{i-1})$. This kind of
models assume independence between the word $w_i$ and the words
not belonging to the $n$-gram, $w_j,\,j<i-n$. This is a drawback
for unbounded dependencies but we are not interested in capturing
the complete grammatical relationships. We intend to capture 
the probabilities of smaller
constructions which may hold positive/negative sentiment.
Another assumption we make is independence between different sentences.

In Opinum the words are lemmas (or wildcards replacing entities),
and the number of words among which we assume dependence is $n=5$.
A maximum $n$ of 5 or 6 is common in machine translation where huge amounts of
text are used for building a language model (\cite{Moses:07}). In our case we have at our
disposal a small amount of data but the language is drastically
simplified by removing the morphology and entities, as previously
explained. We have experimentally found that $n>5$ does not
improve the classification performance of lemmatized opinions and could 
incur over-fitting. 

In our setup we use the IRSTLM open-source library 
for building the language model. It performs an $n$-gram count for all
$n$-grams from $n=1$ to $n=5$ in our case. To deal with data sparseness
a redistribution of the zero-frequency probabilities is performed for
those sets of words which have not been observed in the training set
$\mathcal{L}(\mathbf{O})$. Relative frequencies are discounted to assign
positive probabilities to every possible $n$-gram. Finally a smoothing
method is applied. Details about the process can be found in \cite{IRSTLM:07}.
Another language model approach based on $n$-grams was 
reported in \cite{comparative2006}, where they used the CMU-Cambridge Language Modeling
Toolkit on the original texts.

For Opinum we run IRSTLM twice during the training phase: once taking as input the
opinions labeled as positive and once taking the negatives:
\begin{eqnarray}
\mathit{M^p} &\leftarrow& \text{Irstlm}\left(\mathcal{L}\left(\mathbf{O}^p\right)\right) \nonumber \\
\mathit{M^n} &\leftarrow& \text{Irstlm}\left(\mathcal{L}\left(\mathbf{O}^n\right)\right) \nonumber 
\end{eqnarray}
These two models are further used for querying new opinions on
them and deciding whether it is positive or negative, as detailed
in the next subsection.

\subsection{Evaluation and confidence}

In the Opinum system we query the $\mathit{M^p},\mathit{M^n}$ models with
the \cite{KenLM:2011} KenLM open-source library because it answers the queries
very quickly and has a short loading time, which is suitable
for a web application. It also has an efficient memory
management which is positive for simultaneous queries to the server.

The queries are performed at sentence level. Each sentence $s \in o_t$
is assigned a score which is the log probability of the 
sentence being generated by the language model. The decision is taken
by comparing its scores for the positive and for the negative models.
For a given opinion $o_t$, the log-probability sums can be taken:
$$
d_{o_t}=\sum_{s\in o_t} \log P(s|\mathit{M}^p) - \sum_{s\in o_t} \log P(s|\mathit{M}^n)
\begin{array}{c}
                                          \\
                                         \gtrless \\
                                         ?
\end{array}
0
$$
\newcommand{\sign}{\operatorname{sign}}
If this difference is close to zero, $|d_{o_t}|/w_{o_t}<\varepsilon_0$, it can be considered that the classification
is neutral. The number of words $w_{o_t}$ is used as a normalization
factor. If it is large, $|d_{o_t}|/w_{o_t}>\varepsilon_1$, it can be
considered that the opinion has a very positive or very negative sentiment.
Therefore Opinum classifies the opinions with qualifiers: \textit{very/somewhat/little positive/negative}
depending on the magnitude $|d_{o_t}|/w_{o_t}$ and $\sign(d_{o_t})$, respectively.

The previous assessment is also accompanied by a confidence measure given by the
level of agreement among the different sentences of an opinion. If all its
sentences have the same positivity/negativity, measured by 
$\sign(d_{s_j}),\,s_j\in o$, 
with large magnitudes then
the confidence is
the highest. In the opposite case in which there is the same number of 
positive and negative sentences with similar magnitudes the confidence is the lowest.
The intermediate cases are those with sentences agreeing in sign but some of them
with very low magnitude, and those with most sentences of the same sign and some
with different sign.
We use Shannon's entropy measure $H(\cdot)$ to quantify the amount of disagreement.
For its estimation we divide the range of possible values of $d$ in $B$ ranges, referred to as bins:
$$
H_{o_t} = \displaystyle \sum_{b=1}^B p(d_b)\log\dfrac{1}{p(d_b)}.
$$
The number of bins should be low (less than 10), otherwise it is difficult to
get a low entropy measure because of the sparse values of $d_b$.
We set two thresholds $\eta_0$ and $\eta_1$ such that the confidence
is said to be \textit{high/normal/low} if $H_{o_t}<\eta_0$, $\,\eta_0<H_{o_t}<\eta_1$ or
$H_{o_t}>\eta_1$, respectively

The thresholds $\varepsilon$, $\eta$ and the number of bins $B$ are experimentally 
set. The reason for this is that they are used to tune subjective qualifiers 
(very/little, high/low confidence) and will usually
depend on the training set and on the requirements of the application. 
Note that the classification in positive or negative sentiment
is not affected by these parameters. 
From a human point of view it is also a subjective 
assessment but in our setup it is looked at as a feature implicitly given by the 
labeled opinions of the training set.

\section{Experiments\label{sec:experiments}}

\begin{table*}
  \small
  \begin{tabular}{|l|l|c|} \cline{1-2}
   \multicolumn{2}{|c|}{Similar words, different meaning}& \multicolumn{1}{c}{}\\ \hline
   Original Spanish text & Meaning in English &  Result\\ \hline
   \begin{tabular}{l} ``Al tener la web, no pierdes \\ el tiempo por tel\'efono.''\end{tabular} & 
     \begin{tabular}{l} As you have the website you \\ don't waste time on the phone.\end{tabular} & Positive \\ 
   \begin{tabular}{l} ``En el teléfono os hacen perder\\ el tiempo y no tienen web.''\end{tabular} & 
     \begin{tabular}{l} They waste your time on the phone\\ and they don't have a website.\end{tabular}  & Negative \\ \hline 
   \begin{tabular}{l} ``De todas formas me\\ solucionaron el problema.''\end{tabular} & 
     \begin{tabular}{l} Anyway, they solved my problem.\end{tabular} &  Positive \\ 
   \begin{tabular}{l} ``No hay forma de que\\ me solucionen el problema.''\end{tabular} & 
     \begin{tabular}{l} There is no way to make them \\solve my problem.\end{tabular} &  Negative \\ \hline
  \end{tabular}
  
  \caption{Opinum for financial opinions in Spanish. Short examples of successful classification
  which may be attributed to the $n$-grams models (order $n=5$). These examples can be tested on-line at http://www.prompsit.com/en/opinum, in the 2012 version.\label{tabla1}}
\end{table*}

\begin{table*}
  \small
  \begin{tabular}{|l|l|c|} \cline{1-2}
    \multicolumn{2}{|c|}{A negative opinion with several sentences}& \multicolumn{1}{c}{}\\ \hline
   Original Spanish text & Meaning in English &  Result\\ \hline
   \begin{tabular}{l} ``Con ENTIDAD me fue muy\\ bien.''\end{tabular} & 
     \begin{tabular}{l} I was fine with ENTITY. \end{tabular} & Positive \\ 
   \begin{tabular}{l} ``Hasta que surgieron los\\ problemas.''\end{tabular} & 
     \begin{tabular}{l} Until the problems began. \end{tabular} & Negative \\ 
   \begin{tabular}{l} ``Por hacerme cliente me \\ regalaban 100 euros.''\end{tabular} & 
     \begin{tabular}{l} They gave me 100 euro for\\ becoming a client. \end{tabular} & Positive \\ 
   \begin{tabular}{l} ``Pero una vez que eres cliente\\ no te aportan nada bueno.''\end{tabular} & 
     \begin{tabular}{l} But once you are a client, they\\ they do not offer anything good.\end{tabular} & Negative \\ 
   \begin{tabular}{l} ``Estoy pensando cambiar de\\ banco.''\end{tabular} & 
     \begin{tabular}{l} I am considering switching to\\ another bank. \end{tabular} & Negative \\  \hline
   \multicolumn{2}{|l|}{Classification of the complete opinion}& Negative \\ \hline
  \end{tabular}
 
  \caption{Example of an opinion with several sentences which is classified as Negative.\label{tabla2}}
\end{table*}

In our experimental setup we have a set of positive and negative
opinions in Spanish, collected from a web site for user reviews and opinions.
The opinions are constrained to the financial field including
banks, savings accounts, loans, mortgages, investments, credit cards, and all other
related topics. The authors of the opinions are not professionals, they are
mainly customers. There is no structure required for their opinions, and they
are free to tell their experience, their opinion or their feeling about the
entity or the product. The users meant to communicate their review
to other humans and they don't bear in mind any natural language processing
tools. The authors decide whether their own opinion is positive or
negative and this field is mandatory. 

The users provide a number of stars as well:
from one to five, but we have not used this information. It is interesting to
note that there are 66 opinions with only one star which are marked as positive. 
There are also 67 opinions with five stars which are marked as negative. This
is partially due to human errors, a human can notice when reading them. However
we have not filtered these noisy data, as removing human errors could be regarded
as biasing the data set with our own subjective criteria.

% boyan@s15412536:~/sentiment/opinionclassifier/data/finance19$ 
%find . -name "*.txt" | sed 's/[1-5]0_._/#/g' | sed 's/[0-9]//g' | sort | uniq | wc -l

Regarding the size of the corpus, it consists of 9320 opinions about 180
different Spanish banks and financial products. From these opinions 
5877 are positive and 3443 are negative. There is a total of 709741 words and
the mean length of the opinions is 282 words for the positive and 300 words for the
negative ones. In the experiments we present in this work, we randomly divide
the data set in 75\% for training and 25\% for testing. We check that the
distribution of positive and negative remains the same among test and train.

%boyan@s15412536:~/sentiment/opinionclassifier2/data-gen/finance19V4_lemmastrain$ grep -ro UNKNOWN * | wc -l
After the $\mathcal{L}_2(\cdot)$ lemmatization and entity substitution, the
number of different words in the data set is 13067 in contrast with the 78470 different words
in the original texts. In other words, the lexical complexity is reduced by 83\%.
Different substitutions play a different role in this simplification.
The ``Unknown'' wildcard represents a 7,13\% of the original text. %50638
Entities were detected and replaced 33858 times (7807 locations, 5409 people, 19049 companies,
502 e-mails addresses and phone numbers, 2055 URLs, 1136 dates)
which is a 4,77\% of the text. There are also 46780 number substitutions, a 7\% of the text.
The rest of complexity reduction is due to the removal of the morphology as explained in 
Subsection~\ref{subsec:lemmatization}.

In our experiments, the training of Opinum consisted of lemmatizing and substituting entities
of the 6990 opinions belonging the training set and building the language models.
The positive model is built from 4403 positive opinions and the negative model is built 
from 2587 negative opinions. Balancing the amount of positive and negative
samples does not improve the performance. Instead, it obliges us to remove 
an important amount of positive opinions and the classification results are
decreased by approximately 2\%. This is why we use all the opinions available
in the training set. Both language models are $n$-grams with $n\in [1,5]$.
Having a 37\% less samples for the negative opinions is not a problem thanks to
the smoothing techniques applied by IRSTLM. Nonetheless if the amount of training texts
is too low we would recommend taking a lower $n$. A simple way to set $n$ is
to take the lowest value of $n$ for which classification performance is improved.
Depending on the language model tool used, an unnecessarily high $n$ could overfit the models. 
The effect of the order is further analysed in Section~\ref{sec:discussion}. An example of
the contribution of $n$-grams to successfully capture small order differences is
shown in Table~\ref{tabla1}.

\begin{figure}[thb]
 \centering
 \includegraphics[width=0.7\textwidth]{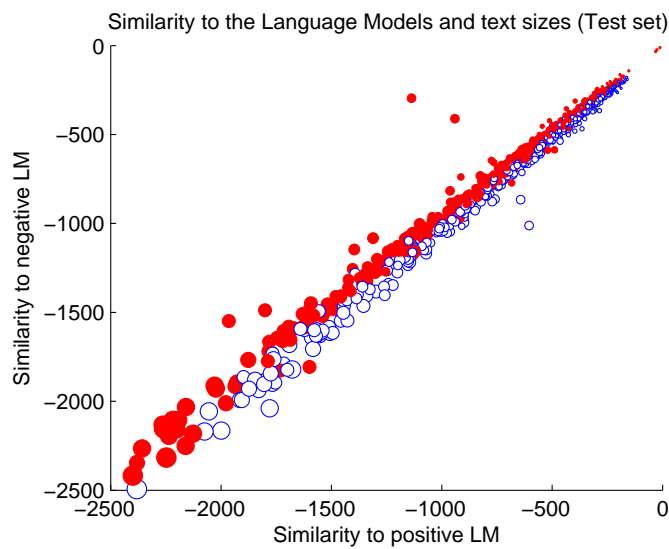}
 \caption{\label{fig-sizepoints}
  Relation between similarity to the models (x and y axis) and the relative size of the opinions
  (size of the points).
 }
\end{figure}

The tests are performed with 2330 opinions (not involved in building the
models). For measuring the accuracy we do not use the qualifiers information
but only the decision about the positive or negative class. In Figure~\ref{fig-sizepoints}
we show the scores of the opinions for the positive and
negative models. The score is the sum of scores of the sentences, thus it 
can be seen that longer opinions (bigger markers) have bigger scores.
Independence of the size is not necessary for classifying in positive and
negative. In the diagonal it can be seen that positive samples are close
to the negative ones, this is to be expected: both positive and negative 
language models are built for
the same language. However the small difference in their scores yields an $81.98$\% 
success rate in the classification. An improvement of this rate would be difficult to
achieve taking into account that there is noise in the training set and
that there are opinions without a clear positive or negative feeling.
A larger corpus would also contribute to a better result. Even though we have
placed many efforts in simplifying the text, this does not help in the cases
in which a construction of words is never found in the corpus. A construction could even be
present in the corpus but in the wrong class. For instance, in our corpus 
``no estoy satisfecho'' (meaning ``I am not satisfied'') appears 3 times among the
positive opinions and 0 times among the negative ones. This weakness of the corpus
is due to sentences referring to a money back guarantee: 
``si no esta satisfecho le devolvemos el dinero'' which are used in a positive context.

Opinum performs slightly better with long opinions, rather than short sentences.
We attribute this to the fact that in long opinions a single sentence does not usually 
change much the positiveness score.
For some examples see Table~\ref{tabla2}.
In long opinions every sentence is prone to show the sentiment except for the cases
of irony or opinions with an objective part.

An installation of Opinum is available on-line. It lacks usability
facilities like batch processing and files uploading because it is installed for small
test purposes. In the web interface  we only provide the single
opinion query and we output the decision, the 
qualifiers information and the confidence measure.
For a better performance, the system can be installed in a Linux-based machine and its scripts
can be used in batch mode.

The query time of Opinum on a standard computer ranges 
from $1.63$ s for the shortest opinions to
$1.67$ s for those with more than 1000 words.
In our setup, most of the time is spent in loading the morphological dictionary,
few milliseconds are spent in the morphological analysis of the opinion
and the named entity substitution, and less than a millisecond is spent
in querying each model. In a batch mode, the morphological analysis could be
done for all the opinions together and thousands of them could be
evaluated in seconds.

% esta figura confunde
% \begin{figure}[thb]
%  \centering
%  \includegraphics[width=0.5\textwidth]{sizedistr}
%  \caption{\label{fig-sizedistr}
%   Relation between similarity to the models (x and y axis) and the size of the text
%   (represented with the size of the points).
%  }
% \end{figure}

\section{Discussion on performance \label{sec:discussion}}

\begin{figure}[thb]
 \centering
 \includegraphics[width=0.7\textwidth]{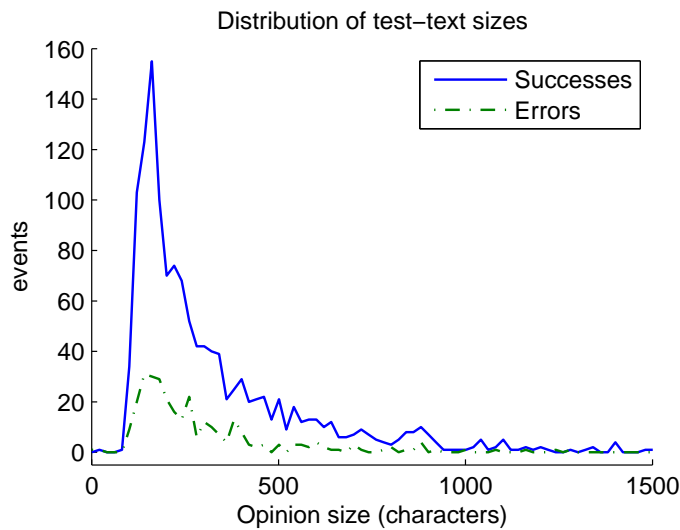}
 \caption{\label{fig-sizedistr}
  Number of successful and erroneous classifications (vertical axis) depending 
  on the size of the test opinions (horizontal axis).
 }
\end{figure}

\begin{figure}[thb]
 \centering
 \includegraphics[width=0.49\textwidth]{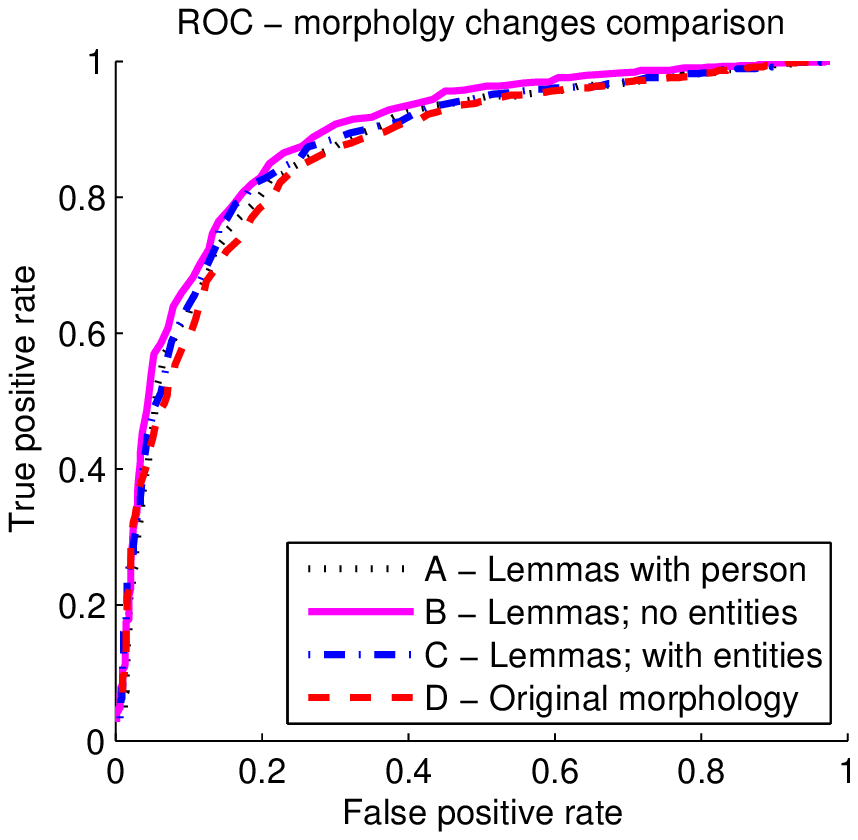}
 \includegraphics[width=0.49\textwidth]{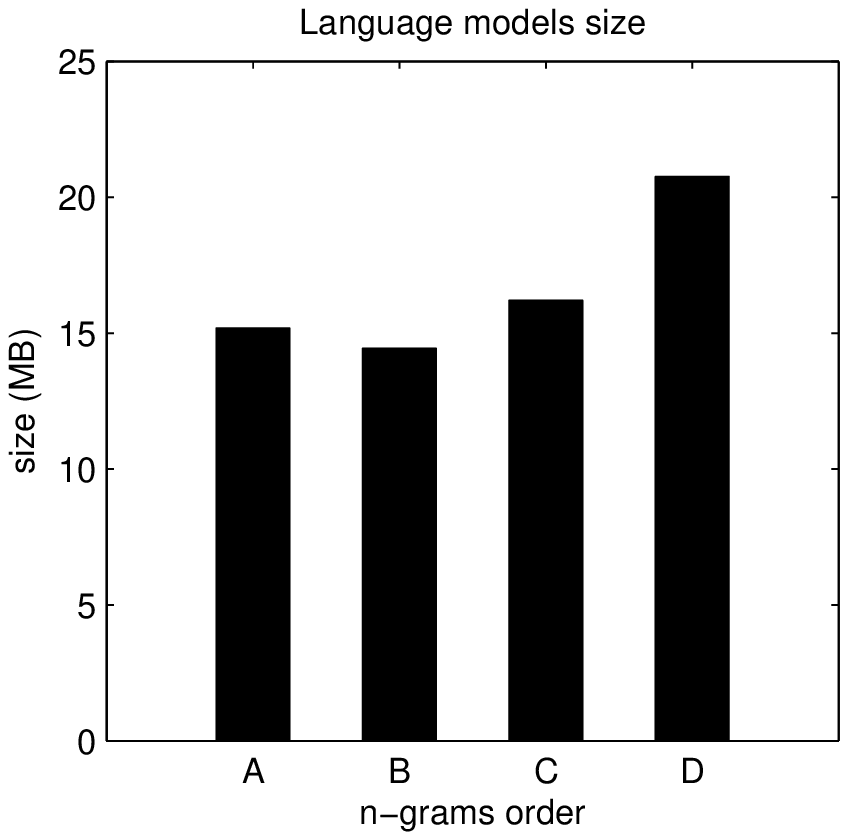}
 \caption{\label{fig-roc-morph}
  Left: Receiver Operating Characteristic (ROC) curves for four language models (A,B,C and D) with different morphological changes, as described in Table~\ref{tab-roc-morph}. Right: The size of these models.
 }
\end{figure}

\begin{figure}[thb]
 \centering
 \includegraphics[width=0.49\textwidth]{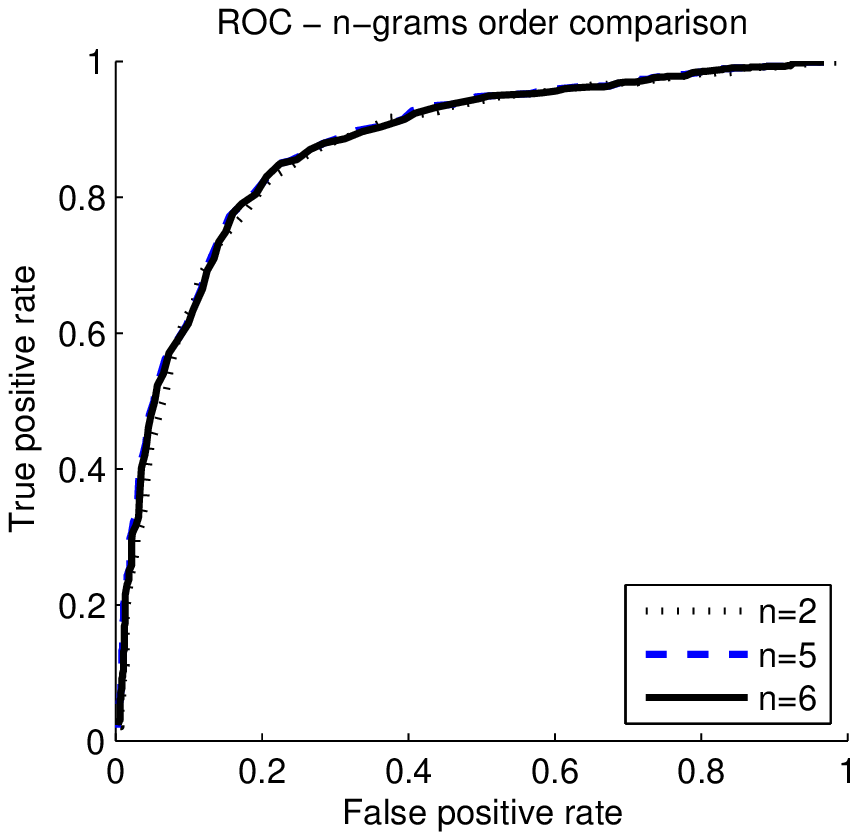}
 \includegraphics[width=0.49\textwidth]{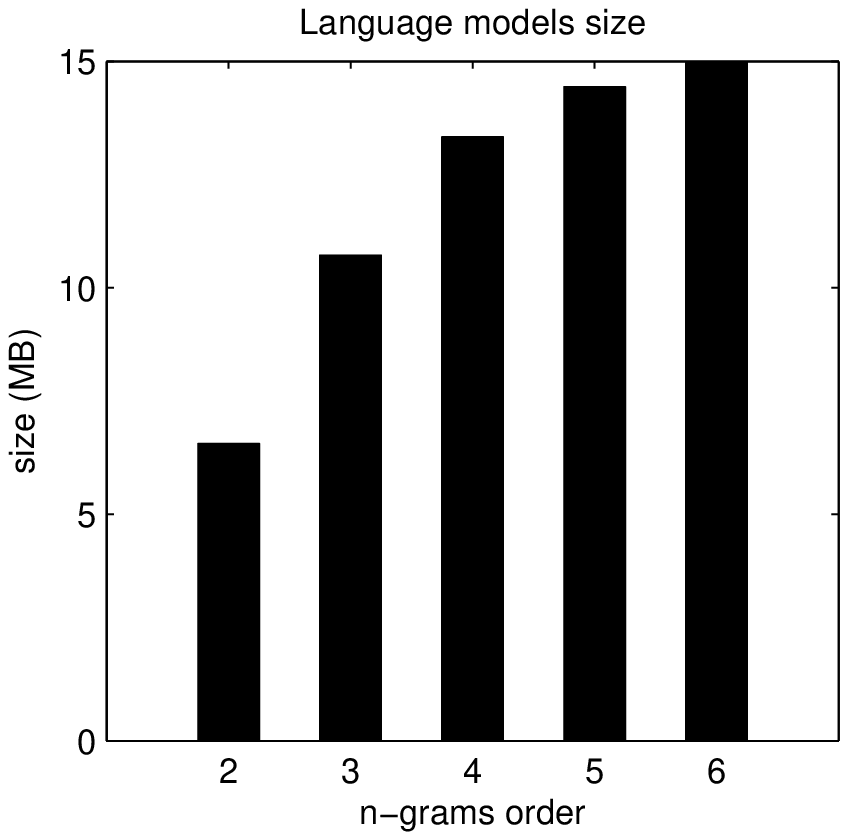}
 \caption{\label{fig-roc-order}
  Left: Receiver Operating Characteristic (ROC) curves for four language models with different
   $n$-grams order. Right: size of the models for different  $n$-grams order. In all of them
   the same morphological changes are used: only lemmas and no named entities.
 }
\end{figure}

\begin{figure}[thb]
 \centering
 \includegraphics[width=0.49\textwidth]{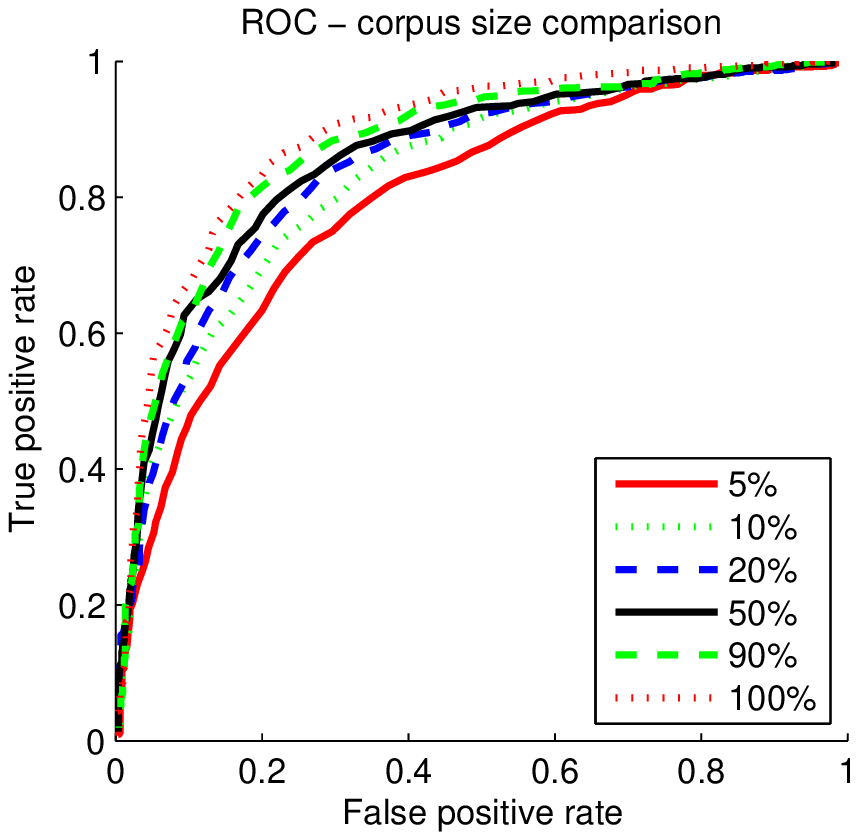}
 \includegraphics[width=0.49\textwidth]{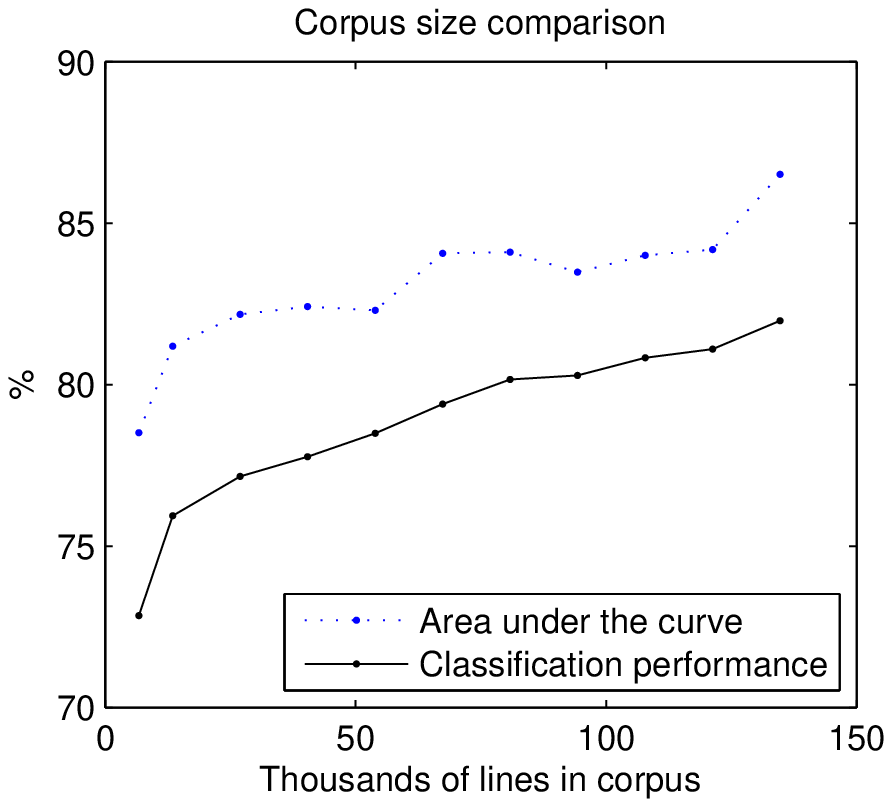}
 \caption{\label{fig-roc-lines}
  Receiver Operating Characteristic (ROC) curve of the Opinum classifier for financial opinions.
 }
\end{figure}

The performance of Opinum depending on the size of the opinions of the test set is
shown in Figure~\ref{fig-sizedistr}.
In Figures~\ref{fig-roc-morph},~\ref{fig-roc-order} and~\ref{fig-roc-lines} 
the ROC curve of the classifier shows its stability
against changing the true-positive versus false-negative rates.
The success rate of Opinum is $81.98\%$, improving significantly the $69\%$ baseline given by
a classifier based on the frequencies of single words, on the same data set.
A comparison with 
other methods would be a valuable source of evaluation. It is not 
feasible at this moment because of the lack of free customers opinions databases
and opionion classifiers as well.
However in the present work we discuss important aspects of performance under
different experimental conditions.

The first of them is the size of the opinions. In Figure~\ref{fig-sizedistr} we show
the relation between successful and erroneous classifications. On the one hand this figure 
gives an idea of the distribution of opinion lengths in the corpus. On the other hand it
shows for which opinions lengths the success rate is better. The performance is similar for
a wide range of opinion lengths: from 2-3 sentences to opinions of several paragraphs.
It can be seen that for unusually long opinions the performance is worse. This can be
attributed to their different style. In middle sized opinions users focus on expressing
a positive or negative feeling in a few sentences, while in longer opinions they are not
so clear. One way to tackle this problem would be to take an approach for detecting the parts 
of the opinion which matter for classification. Another recommendation would be to filter
the unusually short or long opinions from the training set, or even to construct different classifiers
for different ranges of opinion lengths.

% \begin{figure}[thb]
%  \centering
%  \includegraphics[width=0.5\textwidth]{ROC}
%  \caption{\label{fig-roc}
%   Receiver Operating Characteristic (ROC) curve of the Opinum classifier for financial opinions.
%  }
% \end{figure}

Another of the characteristics of Opinum which is worth evaluating is the morphology
substitution approach. In Figure~\ref{fig-roc-morph} we show a comparison of the ROC
curves when using different morphological substitutions. For instance in the classifier
denoted with label A, morphology is removed but the information about 1st, 2nd or 3rd person
is kept. In B morphology is removed and named entities (companies, people, dates, numbers)
are substituted by wildcards. In C only lemmas are used and in D the original text (tokenized)
is used. This information is summarized in Table~\ref{tab-roc-morph}. The performances of
these four classifiers are similar. B outperforms the rest, which means that the proposed 
strategy to keep the person information is not useful. Moreover the language model size for
B is smaller. A possible reason reason for the simpler model outperforming the more complex is
that there is not enough training data. As expected, the model based on the original text
performs worse and is the biggest (the size includes the sum of the positive and negative models).

\begin{table}[thb]
 \begin{center}
 \begin{tabular}{l|c|c|c|c|}
       & A & B & C & D \\      
       \hline
 Morphology removed &\checkmark&\checkmark&\checkmark& \\
 Person not included& &\checkmark&\checkmark& \\
 Unknown words replaced &\checkmark&\checkmark& & \\
 Named entities replaced& \checkmark&\checkmark& & \\
 \hline
 %language model size & $15.18$MB & $14.43$MB & $16.20$MB & $20.76$MB \\
 Classifier success rate&$80.76\%$& $ 81.97\%$ & $81.61\% $ & $79.85\% $ \\
 Area under the curve &$83.44\%$& $86.51\% $ & $83.97\% $ & $81.00\% $ \\
 \hline
 \end{tabular}
 \end{center}
 \caption{\label{tab-roc-morph} Characteristics of the language models of Figure~\ref{fig-roc-morph}. }
\end{table}

Another question that has to be addressed is the maximum order of the $n$-grams of the language model. In statistical machine translation orders between 5 and 7 are common. However in machine translation the corpora used to train the models are considerably larger than our 709741-words corpus of opinions. Due to this there is no significant difference in the performance of higher order models, and even bi-grams perform well (Figure~\ref{fig-roc-order}-Left). The reason for similar performance is that infrequent $n$-grams are pruned.  The actual difference is in the size of the models (Figure~\ref{fig-roc-order}-Right). In our case we select a maximal order 5, because the classification performance is slightly higher. The benefit of a high order would be increased with bigger corpora.

Thus the size of the labeled corpus of opinions is of prime importance. It is also the main limitation of most companies which need opinion classification for their particular context, an so, the first bottleneck to take into account in statistical sentiment analysis. In Figure~\ref{fig-roc-lines}-Left the ROC plots show the quality of the classifiers when trained with smaller subsets of our original corpus. In Figure~\ref{fig-roc-lines}-Right we show how the classification performance and the area under the curve increase with increasing size of corpus. In these figures the maximal amount of $100\%$ and the maximal amount of $140,000$ lines/paragraphs correspond to the $75\%$ of our corpus because a $25\%$ of it is kept for performance evaluation.

Although in Figure~\ref{fig-roc-lines}-Right the performance exhibits a linear-like growth, this only means that the training set is insufficient for this setup. Of course, language models have a limit for capturing the positive or negative character of a text. In \cite{comparative2006} the authors evaluate opinion classification based on $n$-grams with a larger dataset consisting of $159,558$ positive and
$27,366$ negative opinions. Their results are slightly superior but the data sets are not comparable. They do not present an evaluation for increasing size of the corpus. The fact that their number of positive and negative opinions are very different is another question to address. In our setup we have a still differing but better balanced amount of both classes. We found no improvement by removing the surplus, on the contrary, the performance slightly decreased.

\section{Conclusions and future work\label{sec:conclusions}}

Opinum is a sentiment analysis system designed for classifying
customer opinions in positive and negative. Its approach based on
morphological simplification, entity substitution and $n$-gram 
language models, makes it easily adaptable to other classification
targets different from positive/negative. In this work we present
experiments for Spanish in the financial domain but Opinum could
easily be trained for a different language or domain. To this end an 
Apertium morphological analyser would be necessary (30 languages are
currently available) as well as a labeled data set of opinions.
Setting $n$ for the $n$-gram models depends on the size of the
corpus but it would usually range from 4 to 6, 5 in our case.
There are other parameters which have to be experimentally tuned
and they are not related to the positive or negative classification
but to the subjective qualifier very/somewhat/little and to the
confidence measure. 

The classification performance of Opinum in our financial-domain experiments
is 81,98\% which would be difficult to improve because of the noise
in the data and the subjectivity of the labeling in positive and negative.
The next steps would be to study the possibility to classify in
more than two classes by using several language models. The use of 
an external neutral corpus should also be considered in the future.

In this work we show that one of the most important factors to take
into account is the available number of labeled opinions. The size of
the corpus shows to have much more impact on the performance than the 
order of the $n$-grams or even the morphological simplifications that we
perform. Arguably, morphological simplification would have a greater benefit on larger corpora,
but this has to be further analysed with larger data sets.

In practice the classification performance can be improved by filtering
the corpus from noise. We have kept all the noise and outliers in order to 
be fair with the reality of the data for the present study. However, many
wrong-labeled opinions can be removed. In our data set there were some users
checking the ``recommend'' box and placing one or two stars, or not recommending
the product and giving four or five stars. Also, we have seen that too short opinions
perform insufficiently, and in too long opinions the language does not reflect well the positive or negative
sense. Thus these opinions should be filtered in an application based on Opinum.

As a future work, an important question
is to establish the limitations of this approach for
different domains. Is it equally successful for a wider domain?
For instance, trying to build the models from a mixed set of
opinions of the financial domain and the IT domain. Would it perform well
with a general domain?

Regarding
applications, 
Opinum could be trained for a given domain
without expert knowledge. Its queries are very fast which makes
it feasible for free on-line services.
An interesting application would be to exploit the named entity recognition
and associate positive/negative scores to the entities
based on their surrounding text. If several domains were available,
then the same entities would have different scores depending on the domain,
which would be a valuable analysis.

\bibliographystyle{elsarticle-harv}
\bibliography{opinum}

\begin{thebibliography}{12}
\expandafter\ifx\csname natexlab\endcsname\relax\def\natexlab#1{#1}\fi
\expandafter\ifx\csname url\endcsname\relax
  \def\url#1{\texttt{#1}}\fi
\expandafter\ifx\csname urlprefix\endcsname\relax\def\urlprefix{URL }\fi

\bibitem[{Blair-Goldensohn et~al.(2008)Blair-Goldensohn, Neylon, Hannan, Reis,
  Mcdonald, and Reynar}]{Google:08}
Blair-Goldensohn, S., Neylon, T., Hannan, K., Reis, G.~A., Mcdonald, R.,
  Reynar, J., 2008. Building a sentiment summarizer for local service reviews.
  In: In NLP in the Information Explosion Era, NLPIX2008.

\bibitem[{Blitzer et~al.(2007)Blitzer, Dredze, and Pereira}]{Blitzer2007}
Blitzer, J., Dredze, M., Pereira, F., 2007. Biographies, bollywood, boomboxes
  and blenders: Domain adaptation for sentiment classification. In: In ACL. pp.
  187--205.

\bibitem[{Cui et~al.(2006)Cui, Mittal, and Datar}]{comparative2006}
Cui, H., Mittal, V., Datar, M., 2006. Comparative experiments on sentiment
  classification for online product reviews. In: proceedings of the 21st
  national conference on Artificial intelligence - Volume 2. AAAI'06. AAAI
  Press, pp. 1265--1270.

\bibitem[{Federico and Cettolo(2007)}]{IRSTLM:07}
Federico, M., Cettolo, M., 2007. Efficient handling of n-gram language models
  for statistical machine translation. In: Proceedings of the Second Workshop
  on Statistical Machine Translation. StatMT '07. Association for Computational
  Linguistics, Stroudsburg, PA, USA, pp. 88--95.

\bibitem[{Heafield(2011)}]{KenLM:2011}
Heafield, K., 2011. Kenlm: faster and smaller language model queries. In:
  Proceedings of the Sixth Workshop on Statistical Machine Translation. WMT
  '11. Association for Computational Linguistics, Stroudsburg, PA, USA, pp.
  187--197.

\bibitem[{Hu and Liu(2004)}]{Hu2004}
Hu, M., Liu, B., 2004. Mining and summarizing customer reviews. In: Proceedings
  of the tenth ACM SIGKDD international conference on Knowledge discovery and
  data mining. KDD '04. ACM, New York, NY, USA, pp. 168--177.

\bibitem[{Koehn et~al.(2007)Koehn, Hoang, Birch, Callison-Burch, Federico,
  Bertoldi, Cowan, Shen, Moran, Zens, Dyer, Bojar, Constantin, and
  Herbst}]{Moses:07}
Koehn, P., Hoang, H., Birch, A., Callison-Burch, C., Federico, M., Bertoldi,
  N., Cowan, B., Shen, W., Moran, C., Zens, R., Dyer, C., Bojar, O.,
  Constantin, A., Herbst, E., 2007. Moses: open source toolkit for statistical
  machine translation. In: Proceedings of the 45th Annual Meeting of the ACL on
  Interactive Poster and Demonstration Sessions. ACL '07. Association for
  Computational Linguistics, Stroudsburg, PA, USA, pp. 177--180.

\bibitem[{Mcdonald et~al.(2007)Mcdonald, Hannan, Neylon, Wells, and
  Reynar}]{Mcdonald2007}
Mcdonald, R., Hannan, K., Neylon, T., Wells, M., Reynar, J., 2007. Structured
  models for fine-to-coarse sentiment analysis. In: Proceedings of the 45th
  Annual Meeting of the Association of Computational Linguistics.

\bibitem[{Pang et~al.(2002)Pang, Lee, and Vaithyanathan}]{Pang2002}
Pang, B., Lee, L., Vaithyanathan, S., 2002. Thumbs up? sentiment classification
  using machine learning techniques. In: In proceedings of EMNLP. pp. 79--86.

\bibitem[{Raaijmakers et~al.(2008)Raaijmakers, Truong, and
  Wilson}]{Raaijmakers2008}
Raaijmakers, S., Truong, K.~P., Wilson, T., 2008. Multimodal subjectivity
  analysis of multiparty conversation. In: EMNLP. pp. 466--474.

\bibitem[{Taboada et~al.(2011)Taboada, Brooke, Tofiloski, Voll, and
  Stede}]{Taboada2011}
Taboada, M., Brooke, J., Tofiloski, M., Voll, K., Stede, M., 2011.
  Lexicon-based methods for sentiment analysis. Comput. Linguist. 37, 267--307.

\bibitem[{Tyers et~al.(2010)Tyers, S\'anchez-Mart\'inez, Ortiz-Rojas, and
  Forcada}]{Tyers2010}
Tyers, F.~M., S\'anchez-Mart\'inez, F., Ortiz-Rojas, S., Forcada, M.~L., 2010.
  Free/open-source resources in the apertium platform for machine translation
  research and development. The Prague Bulletin of Mathematical
  Linguistics~(93), 67--–76, iSSN: 0032-6585.

\end{thebibliography}

%% Authors are advised to submit their bibtex database files. They are
%% requested to list a bibtex style file in the manuscript if they do
%% not want to use elsarticle-harv.bst.

%% References without bibTeX database:

% \begin{thebibliography}{00}

%% \bibitem must have one of the following forms:
%%   \bibitem[Jones et al.(1990)]{key}...
%%   \bibitem[Jones et al.(1990)Jones, Baker, and Williams]{key}...
%%   \bibitem[Jones et al., 1990]{key}...
%%   \bibitem[\protect\citeauthoryear{Jones, Baker, and Williams}{Jones
%%       et al.}{1990}]{key}...
%%   \bibitem[\protect\citeauthoryear{Jones et al.}{1990}]{key}...
%%   \bibitem[\protect\astroncite{Jones et al.}{1990}]{key}...
%%   \bibitem[\protect\citename{Jones et al., }1990]{key}...
%%   \harvarditem[Jones et al.]{Jones, Baker, and Williams}{1990}{key}...
%%

% \bibitem[ ()]{}

% \end{thebibliography}

% \section*{Vitae}
% \textbf{Boyan Bonev}
% 
% \textbf{Gema Ram\'irez-S\'anchez}
% 
% \textbf{Sergio Ortiz Rojas}
% 
\end{document}